\documentclass{article}

\usepackage[final]{neurips_2019}

\usepackage[utf8]{inputenc} 
\usepackage[T1]{fontenc}    
\usepackage{hyperref}       
\usepackage{url}            
\usepackage{booktabs}       
\usepackage{amsfonts}       
\usepackage{nicefrac}       
\usepackage{microtype}      
\usepackage{times}
\usepackage{graphicx} 
\usepackage{subcaption}
\usepackage{tabularx}
\usepackage{bbm}
\usepackage{url}
\usepackage{amsmath}
\usepackage{caption}
\usepackage{float}
\usepackage{amssymb}
\usepackage{grffile}
\usepackage{epstopdf}
\usepackage[colorinlistoftodos]{todonotes}

\newcommand{\argmin}[1]{\underset{#1}{\operatorname{argmin}}}

\newcommand{\R}{{\mathbb{R}}}
\newcommand{\E}{{\mathbb{E}}}

\newcommand{\cc}{\lambda}
\newcommand{\F}{\cal{F}}

\newcommand{\MREClog}{MRE-C-$\log$ }

\usepackage{color}

\long\def\pur#1{{\color{red}#1}}
\usepackage[normalem]{ulem}
\long\def\dela#1{{\color{blue}\sout{[#1]}}}

\long\def\oli#1{{\color{green}[#1]}}

\long\def\ss#1{{\color{blue}[#1]}}

\newcommand{\hide}[1]{}

\newtheorem{assumption}{Assumption}

\newtheorem{theorem}{Theorem}
\newtheorem{lemma}{Lemma}
\newtheorem{proposition}{Proposition}

\newtheorem{corollary}{Corollary}

\title{Order Optimal One-Shot Distributed Learning}

\author{Arsalan Sharifnassab,\quad Saber Salehkaleybar,\quad S.~Jamaloddin Golestani \vspace{1mm} \\ 
  Department of Electrical Engineering, 
  Sharif University of Technology, 
  Tehran, Iran\\
  \texttt{a.sharifnassab@gmail.com},\quad \texttt{saleh@sharif.edu},\quad \texttt{golestani@sharif.edu}  \\
}

\begin{document}

\maketitle

\begin{abstract}
We consider distributed statistical optimization in one-shot setting, where there are $m$ machines each observing  $n$ i.i.d. samples. 
Based on its observed samples, each machine then sends an $O(\log(mn))$-length message to a server, at which a parameter minimizing an expected loss is to be estimated.
We propose an algorithm called \emph{Multi-Resolution Estimator} (MRE) whose expected error is no larger than $\tilde{O}\big( m^{-{1}/{\max(d,2)}} n^{-1/2}\big)$, where $d$ is the dimension of the parameter space. This error bound meets existing lower bounds up to poly-logarithmic factors, and is thereby order optimal.
The expected error of MRE, unlike existing algorithms,  tends to zero as the number of machines ($m$) goes to infinity, even when the number of samples per machine ($n$) remains upper bounded by a constant. This property of the MRE algorithm makes it applicable in new machine learning paradigms where $m$ is much larger than $n$.  
\end{abstract}

\section{Introduction}
The rapid growth in the size of datasets has given rise to distributed models for statistical learning, in which   data is not stored on a single machine. 
In several recent learning applications, it is commonplace to distribute data across multiple machines, each of which processes its own data and communicates with other machines to carry out a learning task. 
The main bottleneck in such distributed settings is often the communication between machines, and several recent works have focused on designing communication-efficient algorithms for different machine learning  applications \citep{duchi2012dual,braverman2016communication,chang2017distributed,diakonikolas2017communication,lee2017communication}.

In this paper, we consider the problem of statistical optimization in a distributed setting as follows. 
Consider an unknown distribution $P$ over a collection, $\F$, of differentiable convex functions with Lipschitz first order derivatives, defined on a convex region in $\R^d$. 
There are $m$ machines, each observing $n$ i.i.d sample functions from $P$. Each machine processes its observed data, and transmits a signal of certain length to a server. 
The server then collects all the signals and outputs an estimate of the parameter $\theta^*$ that minimizes the expected loss, i.e., $\min_{\theta}\mathbb{E}_{f\sim P}\big[f(\theta)\big]$.
See Fig.~\ref{Fig:system} for an illustration of the system model.

We focus on the  distributed aspect of the problem considering arbitrarily large number of machines ($m$)  and 
\begin{itemize}
	\item[a)] present an order optimal algorithm with $b=O(\log mn)$ bits per transmission, whose estimation error is no larger than $\tilde{O}\big( m^{-{1}/{\max(d,2)}} n^{-1/2}\big)$,  meeting the lower bound in \citep{saleh2019} up to a poly-logarithmic factor (cf. Theorem~\ref{th:main upper});
	\item[b)] we present an algorithm with a single bit per message with expected error no larger than $\tilde{O}\big(m^{-1/2} + n^{-1/2}  \big)$ (cf. Proposition~\ref{prop:H2constb}).
\end{itemize}

\subsection{Background}
The distributed setting considered here has recently employed in   a new machine learning paradigm called \emph{Federated Learning} \citep{konevcny2015federated}. 
In this framework, training data is kept in users' computing devices due to privacy concerns, and the users participate in the training  process without revealing their data. 
As an example, Google has been working on this paradigm in their recent project, \emph{Gboard} \citep{mcmahan2017federated}, the Google keyboard. 
Besides communication constraints, one of the main challenges in this paradigm is that each machine has a small amount of data. In other words, the system operates in a regime that $m$ is much larger than $n$ \citep{chen2017distributed}.

A large body of distributed statistical optimization/estimation literature considers ``one-shot" setting, in which each machine communicates with the server merely once 
\citep{zhang2013information}. 
In these works, the main objective is to  minimize the number of transmitted bits, while keeping the estimation error as low as the error of a centralized estimator, in which the entire data is co-located in the server.  

If we impose no limit on the communication budget, then each machine can encode its entire data into a single message and sent it to the server. 
In this case, the sever acquires the entire data from all machines, and the distributed problem reduces to a centralized problem.
We call the sum of observed functions at all machines as the centralized empirical loss, and refer to its minimizer as the centralized solution.
It is part of the folklore that the centralized solution is order optimal and its expected error is $\Theta\big(1/\sqrt{mn}\big)$ \citep{lehmann2006theory,zhang2013information}. 
Clearly, no algorithm can beat the performance of the best centralized estimator. 

\citet{zhang2012communication} studied a simple averaging method where each machine obtains the empirical minimizer of its observed functions and sends this minimizer to the server through an $O(\log mn)$ bit message. Output of the server is then the average of all received  empirical minimizers.
\citet{zhang2012communication} showed that the expected error of this algorithm is no larger than $O\big(1/\sqrt{mn}\,+\, 1/n\big)$, provided that:  1- all functions are convex and twice differentiable with Lipschitz continuous second derivatives, and 2- the objective function $\mathbb{E}_{f\sim P}\big[f(\theta)\big]$ is strongly convex at $\theta^*$. 
Under the extra assumption that the functions are three times differentiable with Lipschitz continuous third derivatives, \citet{zhang2012communication} also present a bootstrap method  
whose expected error is $O\big(1/\sqrt{mn}\,+\, 1/n^{1.5}\big)$. 
It is easy to see that, under the above assumptions, the averaging method and the bootstrap method achieve the performance of the centralized solution if $m\leq n$ and $m\leq n^2$, respectively. 
Recently, \citet{jordan2018communication} proposed to optimize a surrogate loss function using Taylor series expansion. This expansion can be constructed at the server by communicating $O(m)$ number of $d$-dimensional vectors. 
Under similar assumption on the loss function as in \citep{zhang2012communication},  they showed that the expected error of their method is no larger than $O\big(1/\sqrt{mn}+1/n^{{9}/{4}}\big)$. 
It, therefore, achieves the performance of the centralized solution for $m \le n^{3.5}$. 
However, note that when $n$ is fixed, 	 all aforementioned bounds remain lower bounded by a positive constant, even when $m$ goes to infinity. 

For the problem of sparse linear regression, \cite{braverman2016communication} proved that any algorithm that achieves optimal minimax squared error, requires to communicate  $\Omega(m\times \min(n,d))$ bits  in total from machines 
to the server. 
Later, \cite{lee2017communication} proposed an algorithm that achieves optimal mean squared error  for the problem of sparse linear regression when $d<n$. 

Recently, \citet{saleh2019} studied  the impact of communication constraints on the expected error, over a class of first order differentiable functions with Lipschitz continuous derivatives. 
In parts of their results, they showed that under the assumptions of Section~\ref{sec:problem def} of this paper in the case of $\log mn$ bits communication budget, the expected error of any estimator is lower bounded by  $\tilde{\Omega}\big( m^{-{1}/{\max(d,2)}} n^{-1/2}  \big)$.
They also showed that if the number of bits per message is bounded by a constant and $n$ is fixed, then the expected error remains lower bounded by a constant, even when the number of machines goes to infinity.

Other than one-shot communication, there is another major communication model that allows for several transmissions back and forth between the machines and the server.  Most existing works of this type \citep{bottou2010large,lian2015asynchronous,zhang2015deep,mcmahan2017communication} 
involve variants of stochastic gradient descent, in which the server queries at each iteration the gradient of empirical loss at certain points from the machines.  
The gradient vectors are then aggregated in the server to update the model's parameters. The expected error of such algorithms typically scales as  $O\big({1}/{k}\big)$, where $k$ is the number of iterations.


\subsection{Our contributions}
We study the problem of one-shot distributed learning under milder assumptions than previously available in the literature.
We assume that loss functions, $f\in\F$, are convex and differentiable with Lipschitz continuous first order derivatives.
This is in contrast to the works of \citep{zhang2012communication} and \citep{jordan2018communication} that assume Lipschitz continuity of second or third derivatives. 
The reader should have in mind this model differences, when comparing our bounds with the existing results.

Unlike existing works, our results concern the regime where the number of machines $m$ is large, and our bounds tend to zero as $m$ goes to infinity, even if the number of per-machine observations $n$ is bounded by a constant.
This is contrary to the algorithms in \citep{zhang2012communication}, whose errors tend to zero only when $n$ goes to infinity. 
In fact, when $n=1$, a simple example\footnote{Consider two convex functions $ f_0(\theta)=\theta^2+\theta^3/6$ and 
	$ f_1(\theta)=(\theta-1)^2+(\theta-1)^3/6$ over $[0,1]$. 
	Consider a  distribution $P$ that associates probability $1/2$ to each function.
	Then, $\mathbb{E}_P[f(\theta)]=f_0(\theta)/2+f_1(\theta)/2$, and the optimal solution is $\theta^*=(\sqrt{15}-3)/2\approx 0.436$. 
	On the other hand, in the averaging method proposed in \citep{zhang2012communication}, assuming $n=1$, 
	the empirical minimizer of each machine is either $0$ if it observes $f_0$, or $1$ if it observes $f_1$.
	Therefore, the server receives messages $0$ and $1$  with equal probability , and $\E\big[\hat{\theta}\big]=1/2$.
	Hence, $\mathbb{E}\big[|\hat{\theta}-\theta^*|\big]>0.06$,  for all values of $m$.}
shows that the expected errors of the simple averaging and bootstrap algorithms in \citep{zhang2012communication} remain lower bounded by a constant, for all values of $m$. The algorithm in \citep{jordan2018communication} suffers from  the same problem and its expected error may not go to zero  when $n=1$.

In this work, we present an algorithm with $O\big(\log(mn)\big)$ bits per message, which we call Multi-Resolution Estimator for Convex landscapes and $\log mn$ bits communication budget (MRE-C-$\log$) algorithm. We show that the estimation error of \MREClog algorithm meets the aforementioned lower bound up to a poly-logarithmic factor. 
More specifically, we prove that the  expected error of \MREClog algorithm is no larger than $O\big( m^{-{1}/{\max(d,2)}} n^{-1/2}  \big)$. 
In this algorithm, each machines reports not only its empirical minimizer, but also some information about the derivative of its empirical loss at some randomly chosen point in a neighborhood of this minimizer.
To provide insight into the underlying idea behind \MREClog algorithm, 
we also present a simple naive approach whose error tends to zero as the number of machines goes to infinity. 
Comparing with the lower bound in \citep{saleh2019}, the expected error of \MREClog algorithm meets the lower bound up to a poly-logarithmic factor.  Moreover, for the case of having constant bits per message, we present a simple algorithm whose error goes to zero with rate  $\tilde{O}\big(m^{-1/2} + n^{-1/2}  \big)$, when $m$ and $n$ go to infinity simultaneously.
We evaluate performance of the \MREClog algorithm in two different machine learning tasks and compare with the existing methods in \citep{zhang2012communication}.  
We show via experiments, for the $n=1$ regime, that \MREClog algorithm outperforms these algorithms. The observations are also in line with the expected error bounds we give in this paper and those previously available. In particular, in the $n=1$ regime, the expected error  of \MREClog 
algorithm goes to zero as the number of machines increases, while the expected errors of the previously available estimators remain lower bounded by a  constant.


\subsection{Outline}
The paper is organized as follows. 
We begin with a detailed model and problem definition in Section~\ref{sec:problem def}.
In Section~\ref{sec:main upper}, we present our algorithms and main upper bounds.
We then report our numerical experiments in Section~\ref{sec:numerical}.
Finally, in Section~\ref{sec:discussion} we discuss our results and present open problems and directions for future  research. 
The proofs of the main results and optimality of the \MREClog algorithm are given in the appendix.


\section{Problem Definition} \label{sec:problem def}
\begin{figure}[t]
	\centering
	\includegraphics[width=6.5cm]{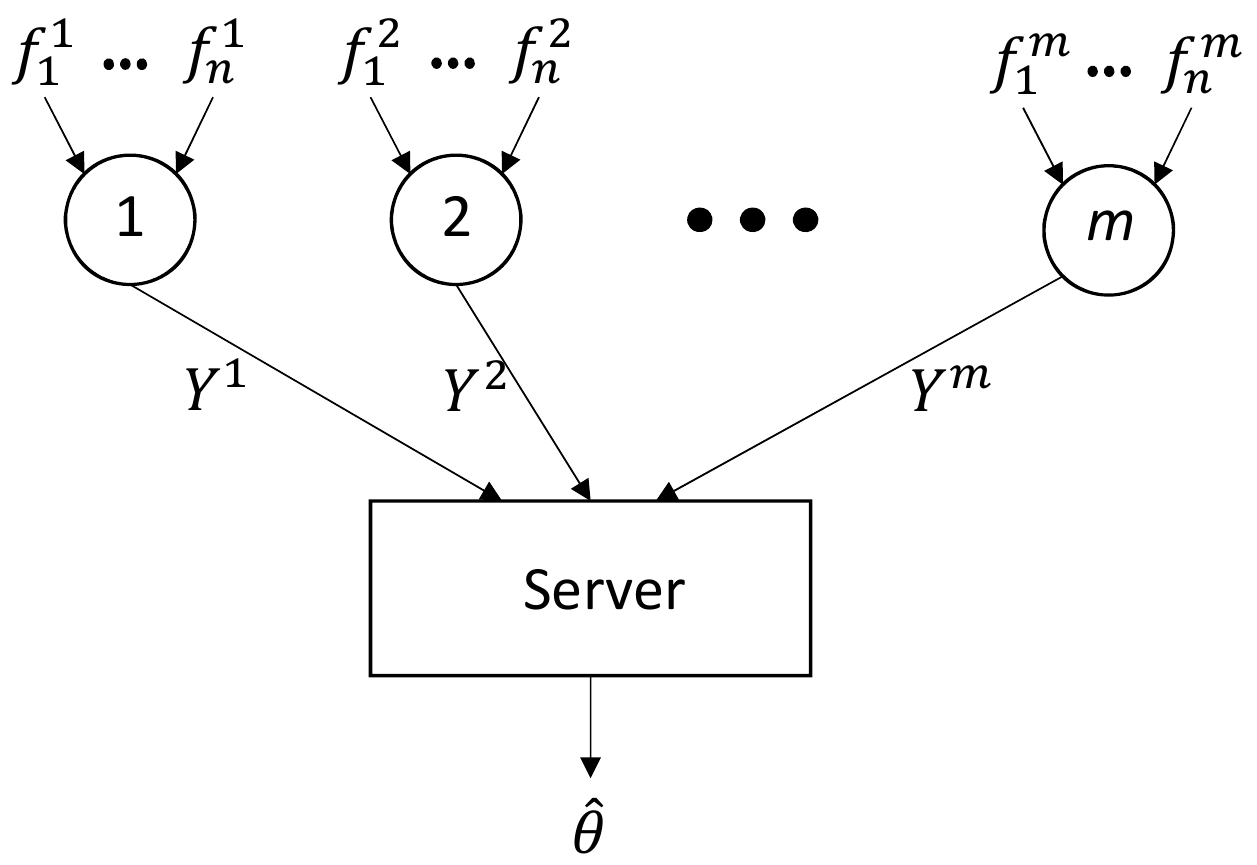}
	\caption{
		A distributed system of $m$ machines, each having access to $n$ independent sample functions from an unknown distribution $P$. 
		Each machine sends a signal to a server based on its observations. The server receives all signals  and output an estimate $\hat{\theta}$ for the optimization problem in \eqref{eq:opt problem}. }
	\label{Fig:system}
\end{figure}

Consider a positive integer $d$ and a  collection  $\mathcal{F}$ of real-valued convex functions over $[-1,1]^d$. 
Let $P$ be an unknown probability distribution over the functions in $\mathcal{F}$.  
Consider the expected loss function
\begin{equation} \label{eq:def of the loss F}
F(\theta) = \mathbb{E}_{f\sim P}\big[f(\theta)\big], \qquad \theta\in [-1,1]^d.
\end{equation} 
Our goal is to learn a parameter $\theta^*$ that minimizes $F$:
\begin{equation}\label{eq:opt problem}
\theta^*=\argmin{\theta \in [-1,1]^d}\, F(\theta) .
\end{equation}
\hide{
	\begin{equation}
	\theta^*=\arg\min_{\theta \in [-1,1]^d}\mathbb{E}_{\dela{x} \pur{f}\sim P}[f(\theta\dela{,x})].
	\end{equation}}
The expected loss is to be minimized in a distributed fashion, as follows.
We consider a distributed system comprising $m$ identical  machines and a server.  Each machine $i$ has access to a set of $n$ independently and identically distributed samples $\{f^i_1,\cdots,f^i_n\}$ drawn from the probability distribution $P$. 
Based on these observed functions,   machine $i$ then sends a signal $Y^i$ to the server.  
We assume that the length of each signal is limited to $b$ bits. 
The server then collects signals  $Y^1,\ldots, Y^m$ and outputs an estimation of $\theta^*$, which we denote by $\hat{\theta}$. 
See Fig.~\ref{Fig:system} for an illustration of the  system model.\footnote{The considered model here is similar to the one in \citep{saleh2019}.} 

\begin{assumption}\label{ass:1}
	We let the following assumptions on $\F$ and $P$ be in effect throughout the paper.
	\begin{itemize}
		\item Every $f\in \F$	 is once differentiable and convex.
		\item  Each $f\in\F$ has bounded and Lipschitz continuous derivatives. More concretely, for any $f\in\F$ and any $\theta,\theta'\in [-1,1]^d$, we have $|f(\theta)|\le \sqrt{d}$,  $\|\nabla f(\theta)\|\leq 1$, and $\|\nabla f(\theta) - \nabla f(\theta')\|\leq \|\theta-\theta'\|$.
		\item  Distribution $P$ is such that $F$ (defined in \eqref{eq:def of the loss F}) is strongly convex. More specifically, there is a constant $\cc>0$ such that for any   $\theta_1,\theta_2 \in [-1,1]^d$, we have $F(\theta_2) \ge F(\theta_1) + \nabla F(\theta_1)^T (\theta_2-\theta_1) + \cc \|\theta_2-\theta_1\|^2$. 
		\item The minimizer of $F$ lies in the interior of the cube $[-1,1]^d$. Equivalently, there exists $\theta^*\in (-1,1)^d$  such that $\nabla F(\theta^*) = \mathbf{0} $.  
	\end{itemize}
\end{assumption}

\section{Algorithms and Main Results} \label{sec:main upper}
In this section, we propose estimators to minimize the expected loss, organized in a sequence of three subsections. In the first subsection, we consider the case  of constant bits per signal transmission, whereas 
in the last two subsections we allow for $\log mn$ bits per  signal transmission. For the latter regime, we first present in Subsection~\ref{sec:naive}, a simple naive approach whose estimation error goes to zero for large values of $m$, even when $n=1$. 
Afterwards, in Subsection~\ref{subsec:main alg}, we describe our main estimator, establish an upper bound on its estimation error, and show that it is order optimal.

\subsection{Constant number of bits per transmission} \label{subsec:constant bit}
Here, we consider a simple case with a one-dimensional domain ($d=1$) and one-bit signal per transmission ($b=1$). 
We show that the expected error can be made arbitrarily small as $m$ and $n$ go to infinity simultaneously.
\begin{proposition}
	Suppose that $d=1$ and $b=1$.
	There exists a randomized estimator $\hat{\theta}$ such that
	\begin{equation*}
	\mathbb{E}\big[(\hat{\theta}-\theta^*)^2\big]^{1/2}=O\left(\frac{1}{\sqrt{n}}+\frac{1}{\sqrt{m}}\right).
	\end{equation*}
	\label{prop:H2constb}
\end{proposition}
The proof is  given in Appendix~\ref{app:constant bit upper bound}. There, we assume for simplicity that the domain is the $[0,1]$ interval and propose a simple randomized algorithm in which each machine $i$ first computes an $O(1/\sqrt{n})$-accurate estimation $\theta^i$ based on its observed functions. 
It then sends a $Y^i=1$ signal with probability $\theta^i$. 
The server then outputs the average of the received signals as the finial estimate.

Based on Proposition~\ref{prop:H2constb}, there is an algorithm that achieves any desired accuracy even with budget of one bit, provided that $m$ and $n$ go to infinity simultaneously. 
In contrary, it was shown in Proposition 1 of \citep{saleh2019} that no estimator yields error better than a constant if $n=1$ and the number of bits per transmission is a constant independent of $m$. 
 We conjecture that the bound in Proposition~\ref{prop:H2constb} is tight. 
More concretely, for constant number of bits per transmission 
and any randomized estimator $\hat{\theta}$, we have $\mathbb{E}[(\hat{\theta}-\theta^*)^2]^{1/2}=\tilde\Omega\big(1/\sqrt{n}+1/\sqrt{m}\big)$.

\subsection{A simple naive approach with $\log mn$ bits per transmission} \label{sec:naive}
We now consider the case where the number of bits per transmission is $O(\log m)$. 
In order to set the stage for our main algorithm given in the next subsection, here we present a simple algorithm and show that its estimation error decays as $O(m^{-1/3})$.
The underlying idea is that unlike existing estimators,  in this algorithm each machine encodes in its signal some information about the shape of its observed functions at a point that is not necessarily close to its own private optimum.
To simplify the presentation, here we confine our setting to one dimensional domain ($d=1$) with each machine observing a single sample function ($n=1$). 
The algorithm is as follows:

\begin{itemize}
	\item[ ]
	Consider a regular grid of size $\sqrt[3]{m}/\log(m)$ over the $[-1,1]$ interval. 
	Each machine $i$ selects a grid point $\theta^i$ uniformly at random. 
	The machine then forms a signal comprising two parts: 1- The location of  $\theta^i$, and 2- The derivative of its observed function $f^i$ at $\theta^i$. 
	In other words, the signal $Y^i$ of the $i$-th machine is an ordered pair of the form $\big(\theta^i,f^{'i}(\theta^i)\big)$, where $f^{'i}(\theta^i)$ is the derivative of $f^i$ at $\theta^i$. 
	In this encoding, we use  $O(\log m)$ bits to represent both $\theta^i$ and $f^{'i}(\theta^i)$.
	In the server, for each grid point $\theta$, the average of $f^{'i}$ is computed over all machines $i$ with $\theta^i=\theta$. 
	We denote this average by $\hat{F}'(\theta)$. The server then outputs a point $\theta$ that minimizes $\big|\hat{F}'(\theta)\big|$. 
\end{itemize}

This algorithm learns an estimation of  derivatives of  $F$, and finds a point that minimizes the size of this derivative.
The following lemma shows that the estimation error of this algorithm is  $\tilde{O}(1/\sqrt[3]{m})$.
The proof is given in Appendix~\ref{app:proof naive}.

\begin{proposition} \label{lem:naive}
	Let  $\hat{\theta}$ be the output of the above estimator. For any $\alpha>1$,
	\begin{equation*}
	\Pr\left(\big|\hat{\theta}-\theta^*\big|>\frac{3\alpha \log(m)}{\lambda \sqrt[3]{m}}\right) \,=\,O\Big(\exp\big(-\alpha^2\log^3m\big)\Big).
	\end{equation*}
	Consequently, for any $k\ge1$, we have $\mathbb{E}\big[|\hat{\theta}-\theta^*|^k\big]=O\big((\log(m)/\sqrt[3]{m})^k\big)$.
\end{proposition}

We now turn to the general case with arbitrary  values for $d$ and $n$, and present our main estimator.

\subsection{The Main Algorithm 
}\label{subsec:main alg}
\begin{figure}[t]
	\centering
	\includegraphics[width=8cm]{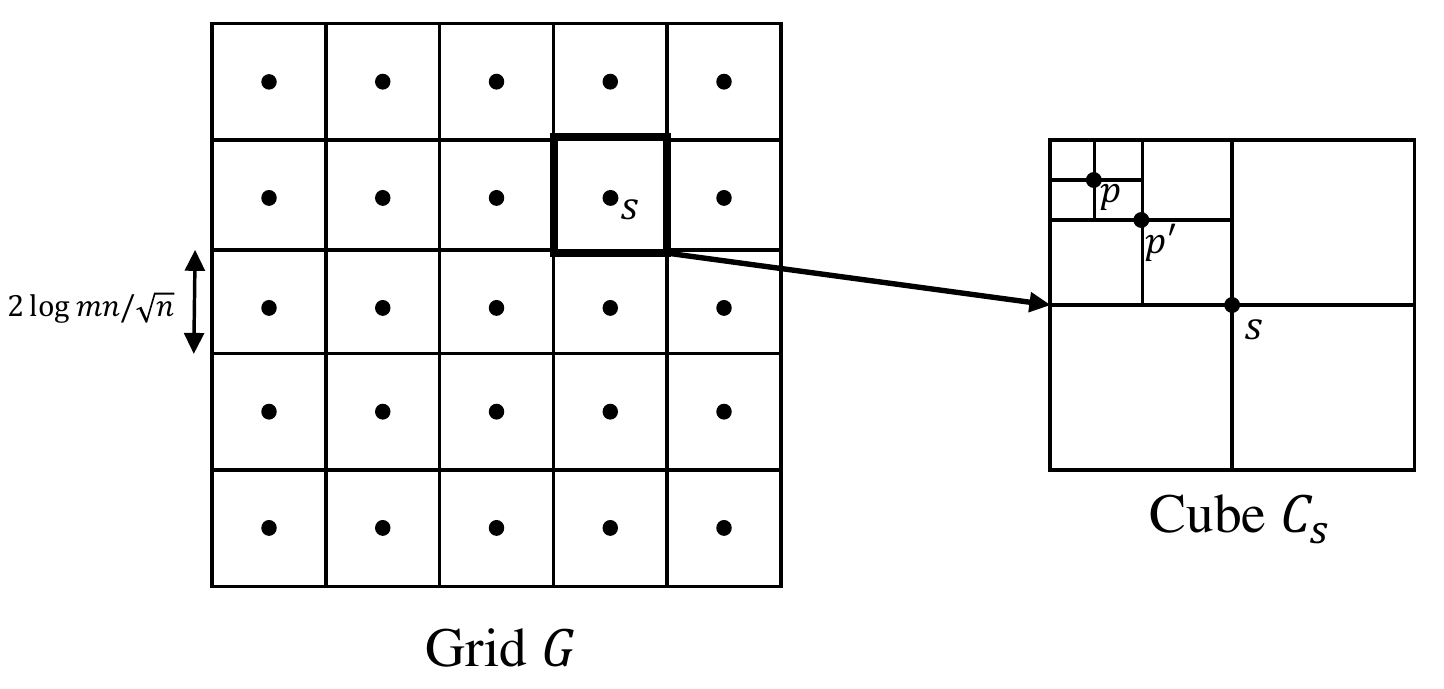}
	\caption{An illustration of grid $G$ and cube $C_s$ centered at point $s$ for $d=2$. The point $p$ belongs to $\tilde{G}_s^2$ and $p'$ is the parent of $p$.}
	\label{Fig:MRE}
\end{figure}

In this part, we propose our main algorithm and an upper bound on its estimation error. In the proposed algorithm, transmitted signals are designed such that the server can construct a multi-resolution view of gradient of function $F(\theta)$ around a promising grid point. Then, we call the proposed algorithm ``Multi-Resolution Estimator for Convex landscapes with $\log mn$ bits communication budget (\MREClog\!)".  The description of \MREClog is as follows:

Each machine $i$ observes $n$ functions and sends a signal $Y^i$ comprising three parts of the form $(s,p,\Delta)$. 
The signals are of length $O(\log(mn))$ bits and the three parts $s$, $p$, and $\Delta$ are as follows.
\begin{itemize}
	\item Part $s$: Consider a grid $G$ with resolution $\log(mn)/\sqrt{n}$ over the $d$-dimensional cube. Each machine $i$ computes the minimizer of the average of its first $n/2$ observed functions,
	\begin{equation} \label{eq:def theta i half upper}
	\theta^i=\argmin{\theta \in [-1,1]^d} \,\sum_{j=1}^{n/2} f_j^i(\theta).
	\end{equation}
	It then lets 	$s$ be the closest grid point to $\theta^i$. 
	\hide{I will apply the following modification: Use $n/2$ of the functions for the $s$ part, and keep the remaining $n/2$ of the functions for the $\Delta$ part. In other words, 
		\begin{equation*}
		\theta^i=\argmin{\theta \in \Theta} \sum_{j=1}^{n/2} f_j^i(\theta).
		\end{equation*}
		This change in necessary because in the proofs we need independence between $s$ and $\Delta$.
	}
	\item Part $p$: Let
	\begin{equation}\label{eq:def delta}
	\delta \,\triangleq \,  4\sqrt{d} \left( \frac{\log^5(mn)}{m} \right)^{\frac1{\max(d,2)}}.
	\end{equation}
	Note that $\delta = \tilde{O}\big(m^{-1/\max(d,2)}\big)$.
	Let $t = \log(1/\delta)$. 
	Without loss of generality we assume that $t$ is an integer.
	Let $C_s$  be a $d$-dimensional cube with edge size $2\log(mn)/\sqrt{n}$ centered at $s$.
	Consider a sequence of $t+1$ grids on $C_s$ as follows.
	For each $l=0,\ldots,t$, we partition the cube $C_s$ into $2^{ld}$ smaller equal sub-cubes with edge size $2^{-l+1} \log(mn)/\sqrt{n}$. 
	The $l$th grid $\tilde{G}_s^l$ comprises the centers of these smaller cubes.
	Then, each $\tilde{G}_s^l$ has $2^{ld}$ grid points.
	For any point  $p'$ in $\tilde{G}_s^l$, we say that $p'$ is the parent of all $2^d$ points in $\tilde{G}_s^{l+1}$
	that are in the $\big(2^{-l}\times (2\log mn)/\sqrt{n}\big)$-cube centered at $p'$ (see Fig. \ref{Fig:MRE}). 
	Thus, each point $\tilde{G}_s^l$ ($l<t$) has $2^d$ children. 
	
	To select $p$, we randomly choose an $l$ from $0,\dots, t$ with probability $2^{(d-2)l}/(\sum_{j=0}^t 2^{(d-2)j})$. 
	We then let $p$ be a uniformly chosen random grid point in $\tilde{G}_s^l$.
	Note that $O(d\log(1/\delta))=O(d\log(mn))$ bits suffice to identify $p$ uniquely. 
	
	\item Part $\Delta$: We let 
	\begin{equation}\label{eq:def hat F upper}
	\hat{F}^i(\theta)\triangleq \frac2n\sum_{j=n/2+1}^{n} f_j^i(\theta),
	\end{equation}
	and refer to it as the empirical function of the $i$th machine.
	If the selected $p$ in the previous part is in $\tilde{G}_s^0$, i.e., $p=s$, then we set $\Delta$ to the gradient of $\hat{F}^i$  at $\theta=s$.  
	\hide{This will change into $\hat{F}^i(\theta)\triangleq 2/n\sum_{j=n/2+1}^n f_j^i(\theta)$.}
	Otherwise, if $p$ is in $\tilde{G}_s^l$ for $l\geq 1$, we let
	\begin{equation*}
	\Delta \,\triangleq\, \nabla \hat{F}^i (p)-\nabla \hat{F}^i(p'),
	\end{equation*}
	where $p'\in \tilde{G}_s^{l-1}$ is the parent of $p$. 
	Note that $\Delta$ is a $d$-dimensional vector whose entries are in the range  
	$\big(2^{-l}\sqrt{d}\log(mn)/\sqrt{n}\big) \times \big[-1,+1\big]$.
	This is due to the Lipschitz continuity of the derivative of the functions in $\F$ (cf. Assumption~\ref{ass:1})  and the fact that $\|p-p'\|=2^{-l}\sqrt{d}\log(mn)/\sqrt{n}$. 
	Hence, we can use  $O(d\log(mn))$  bits to represent $\Delta$ within accuracy $2\delta\log(mn)/\sqrt{n}$. 
\end{itemize}

At the server, we choose an $s^*\in G$ that has the largest number of occurrences in the received signals. 
Then, base on the signals corresponding to $\tilde{G}_{s^*}^0$,
we approximate the gradient of $F$ at $s^*$ as
\begin{equation*}
\hat{\nabla} F(s^*)=\frac{1}{N_{s^*}}\sum_{\substack{\mbox{\scriptsize{Signals of the form }} \\ Y^i=(s^*,s^*,\Delta)}}\Delta,
\end{equation*}	
where $N_{s^*}$ is the number of signals containing $s^*$ in the part $p$. 
Then, for any point $p\in \tilde{G}_{s^*}^l$ with $l\geq 1$, we compute
\begin{equation}
\hat{\nabla}F(p)=\hat{\nabla}F(p')+\frac{1}{N_p}\sum_{\substack{\mbox{\scriptsize{Signals of the form }} \\ Y^i=(s^*,p,\Delta)}}\Delta,
\label{eq:page17}
\end{equation}
where $N_p$ is the number of signals having point $p$ in their second argument. 
Finally, the sever lets $\hat{\theta}$ be a grid point $p$ in $\tilde{G}_{s^*}^t$  with the smallest $\|\hat{\nabla}F(p)\|$.

In the  \MREClog algorithm the signals are of length $d/(d+1) \log m+d \log n$ bits, which is no larger than $d \log mn$. 
Please refer to Section~\ref{sec:discussion} for discussions on how the \MREClog algorithm can be extended to work under more general communication constraints. 

\begin{theorem}\label{th:main upper}
	Let $\hat{\theta}$ be the output of the above algorithm. Then, 
	\begin{equation*}
	\Pr\left(\|\hat{\theta}-\theta^*\|> \frac{8d\, \log^{\frac{5}{\max(d,2)}+1} (mn)}{\cc\,m^{\frac{1}{\max(d,2)}}n^{\frac{1}{2}}}\right) \,=\, \exp\Big(-\Omega\big(\log^2(mn)\big)\Big).
	\end{equation*}
\end{theorem}
The proof is given in  Appendix~\ref{sec:proof main alg}.
The proof goes by first showing that $s^*$ is a closest grid point of $G$ to $\theta^*$ with high probability. 
We then show that for any $l\le t$ and any $p\in \tilde{G}_{s^*}^l$, the number of received signals corresponding to $p$ is large enough so that the server obtains a good approximation of $\nabla F$ at $p$. 
Once we have a good approximation $\hat\nabla F$ of $\nabla F$ at all points of $\tilde{G}_{s^*}^t$, a point at which $\hat\nabla F$ has the minimum norm lies close to the minimizer of $F$.

\begin{corollary} \label{cor:upper}
	Let $\hat{\theta}$ be the output of the above algorithm. 
	There is a constant $\eta>0$ such that 
	for any $k\in \mathbb{N}$,
	\begin{equation*}
	\mathbb{E}\big[\|\hat{\theta}-\theta^*\|^k\big]\,<\,\eta \,\left(\frac{8d\, \log^{\frac{5}{\max(d,2)}+1} (mn)}{\cc\,m^{\frac{1}{\max(d,2)}}n^{\frac{1}{2}}}\right)^k.
	\end{equation*}
	Moreover, $\eta$ can be chosen arbitrarily close to $1$, for large enough values of $mn$.
\end{corollary}


The upper bound in Theorem~\ref{th:main upper} matches the lower bound in Theorem 2 of \citep{saleh2019} up to a polylogarithmic factor. 
In this view, the \MREClog algorithm has order optimal error.
Moreover, as we show in Appendix~\ref{sec:proof main alg}, in the course of computations, the server  obtains  an approximation $\hat{F}$  of $F$ such that for any $\theta$ in the cube $C_{s^*}$, we have $\|\nabla \hat{F}(\theta) -\nabla F(\theta)\| = \tilde{O}\big( m^{-1/d}n^{-1/2})$. 
Therefore, the server not only finds the minimizer of $F$, but also obtains an approximation of $F$ at all points inside $C_{s^*}$.
In the special case that $n=1$, we have  $C_{s^*} = [-1,1]^d$,  and as a result, the server would acquire an approximation of $F$ over the entire domain.
This observation suggests the following insight: In the extreme distributed case ($n=1$), finding an $O\big( m^{-1/d})$-accurate minimizer of $\nabla F$ is as hard as finding an $O\big( m^{-1/d})$-accurate approximation of $F$ for all points in the domain.



\section{Experiments} \label{sec:numerical}

We evaluated the performance of \MREClog on two learning tasks and compared with the averaging method (AVGM) in \citep{zhang2012communication}. Recall that in AVGM, each machine sends the empirical risk minimizer of its own data to the server and the average of received parameters at the server is returned in the output.

 The first experiment concerns the problem of ridge regression. Here, each sample $(X,Y)$ is generated based on a linear model $Y=X^T\theta^*+E$, where $X$, $E$, and $\theta^*$ are  sampled from $N(\mathbf{0},I_{d\times d})$, $N(0,0.01)$, and uniform distribution over $[0,1]^d$, respectively. We consider square loss function with $l_2$ norm regularization: $f(\theta)=(\theta^TX-Y)^2+0.1 \|\theta\|_2^2$.
In the second experiment, we perform a logistic regression task, considering sample vector $X$ generated according to $N(\mathbf{0},I_{d\times d})$ and labels $Y$ randomly drawn from $\{-1,1\}$ with probability $\Pr(Y=1|X,\theta^*)=1/(1+\exp(-X^T \theta^*))$.
In both experiments, we consider a two dimensional domain ($d=2$) and assumed that each machine has access to one sample ($n=1$).

\begin{figure}[t!]
	\centering
	\begin{subfigure}[t]{0.5\textwidth}
		\centering
		\includegraphics[width=2.3in]{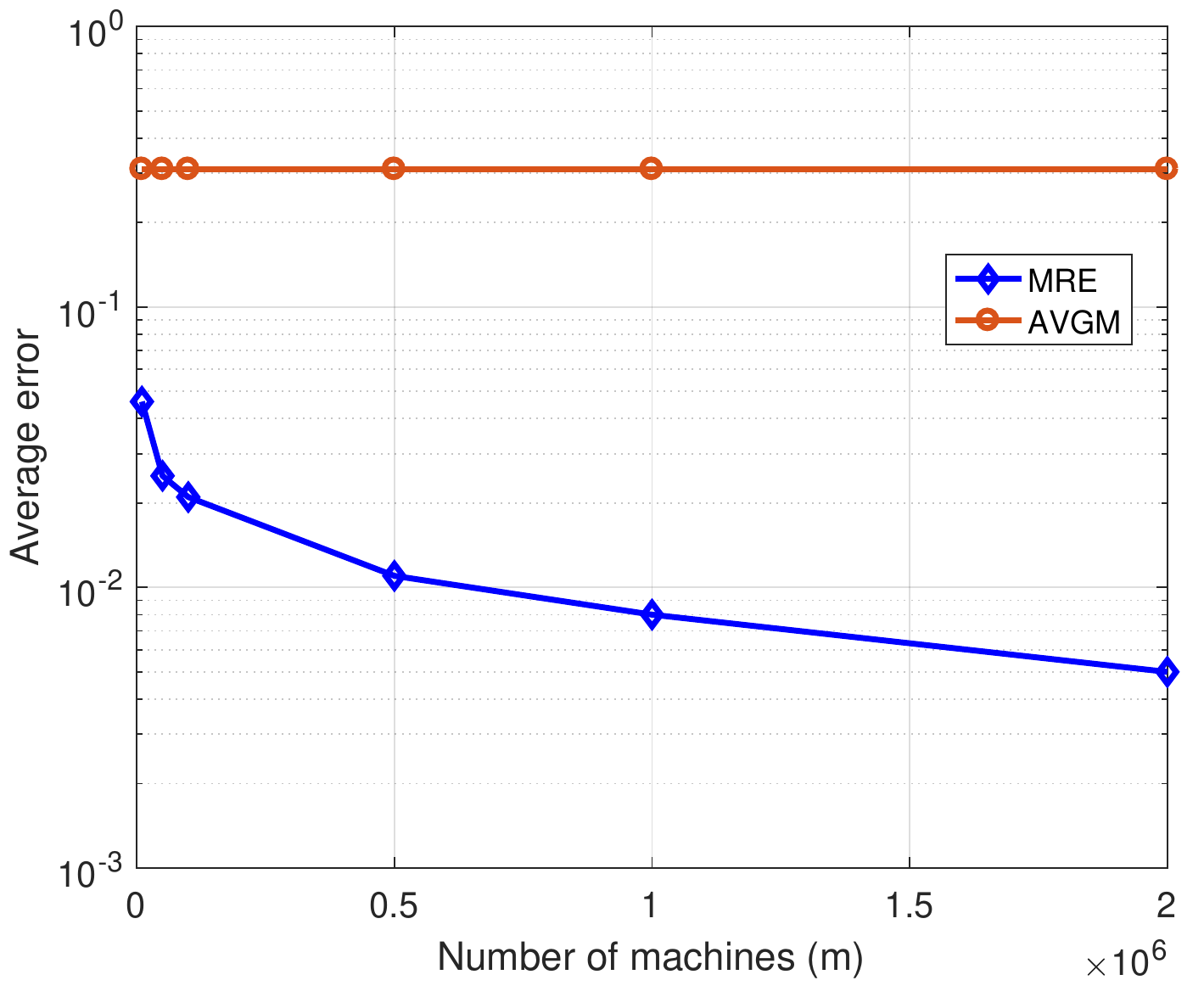}
		\caption{Ridge regression}
	\end{subfigure}%
	~ 
	\begin{subfigure}[t]{0.5\textwidth}
		\centering
		\includegraphics[width=2.35in]{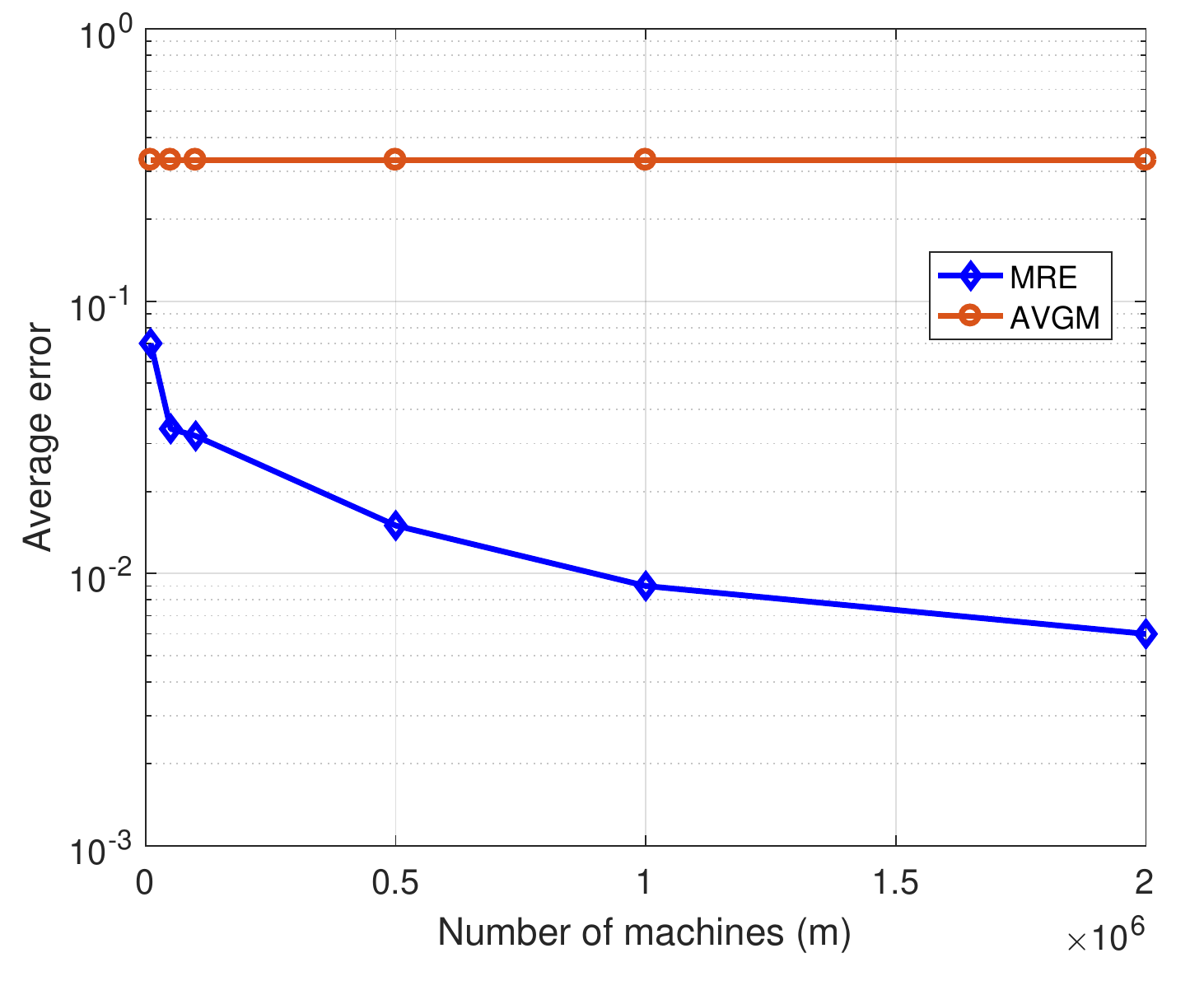}
		\caption{Logistic regression}
	\end{subfigure}
	\caption{The average  of \MREClog and AVGM algorithms versus the number of machines in two different learning tasks. }
	\label{fig:sim}
\end{figure}

In Fig. \ref{fig:sim}, the average of $\|\hat{\theta}-\theta^*\|_2$ is computed over $100$ instances  for the different number of machines in the range $[10^4,10^6]$. Both experiments suggest that the average error of \MREClog keep decreasing as the number of machines increases.  This is  consistent with  the result in Theorem~\ref{th:main upper}, according to which  the expected error of \MREClog is upper bounded by  $\tilde{O}(1/\sqrt{mn})$. 
It is evident from the error curves that \MREClog outperforms the AVGM algorithm in both tasks. This is  because where $m$ is much larger than $n$, the expected error of the  AVGM algorithm typically scales as $O(1/n)$, independent of $m$.


\section{Discussion} \label{sec:discussion}
We studied the problem of statistical optimization in a distributed system with one-shot communications. 
We proposed an algorithm, called \MREClog, with $O\big(\log(mn)\big)$-bits per message, and showed that its expected error is optimal up to a poly-logarithmic factor. 
Aside from being order optimal, the \MREClog algorithm has the advantage over the existing estimators that its error tends to zero as the number of machines goes to infinity, even when the number of samples per machine is upper bounded by a constant.
This property is in line with the out-performance of the \MREClog algorithm in the  $m\gg n$ regime, as discussed in our experimental results. 

The main idea behind the \MREClog algorithm is that it essentially computes, in an efficient way, an approximation of the gradient of the expected loss over the entire domain. 
It then outputs a norm-minimizer of this approximate gradients, as an estimate of the minimizer of the expected loss. 
Therefore, \MREClog carries out the intricate and  seemingly redundant task of approximating the loss function for all points in the domain, in order to resolve the apparently much easier problem of finding a single approximate minimizer for the loss function. 
In this view, it is quite counter-intuitive that such algorithm is order optimal in terms of expected error and sample complexity. 
This observation provides the interesting insight that in a distributed system with one shot communication, finding an approximate minimizer is as hard as finding an approximation of the function derivatives for all points in the domain.

Our algorithms and bounds are designed and derived for a broader class of functions with Lipschitz continuous first order derivatives, compared to the previous works that consider function classes with Lipschitz continuous second or third order derivatives. 
The assumption is indeed both  practically important  and technically challenging. 
For example, it is well-known that the loss landscapes involved in learning applications and neural networks are highly non-smooth. Therefore, relaxing assumptions on higher order derivatives is actually a practically important improvement over the previous works.
On the other hand, considering Lipschitzness only  for the first order derivative  renders the problem way more difficult. To see this, note that when $n>m$, the existing upper bound $O(1/\sqrt{mn}+1/n)$ for the case of Lipschitz second derivatives  goes below the  $O(m^{1/d}n^{1/2})$ lower bound in the  case of Lipschitz first derivatives.

A drawback of the \MREClog algorithm is that each machine requires to know $m$ in order to set the number of levels for the grids. This however can be resolved by considering infinite number of levels, and letting the probability that $p$ is chosen from level $l$ decrease exponentially with $l$. 
Moreover, although communication budget of the \MREClog algorithm is $O(d \log mn)$ bits per signal, the algorithm can be extended to work under more general communication constraints, via dividing each signal to subsignals of length $O(d \log mn)$ each containing an independent independent signal of the \MREClog algorithm. The expected loss of this modified algorithm can be shown to still matches the existing lower bounds up to logarithmic factors. Please refer to \cite{saleh2019} for a thorough treatment.

We also proposed, for $d=1$, an algorithm  with communication budget of one bit per transmission, whose error tends to zero in a rate of $O\big(1/\sqrt{m}+ 1/\sqrt{n}\big)$ as $m$ and $n$ go to infinity simultaneously. We conjecture that this algorithms is order-optimal, in the sense that no randomized constant-bit algorithm has expected error smaller than $O\big(1/\sqrt{m}+ 1/\sqrt{n}\big)$.

There are several open problems and directions for future research. 
The first group of problems involve the constant bit regime. 
It would be interesting if one could verify whether or not the bound in Proposition~\ref{prop:H2constb} is order optimal.
Moreover, the constant bit algorithm in Subsection~\ref{subsec:constant bit} is designed for one-dimensional domains and one-bit per transmission. 
Decent extensions of this algorithm to higher dimensions with vanishing errors under one bit per transmission constraint seem to be non-trivial. 
Investigating the power of more bits per transmission (constants larger than one bit) in reducing the expected error is another interesting direction.

Another important group of problems concerns the more restricted class of functions with Lipschitz continuous second order derivatives. 
Despite several attempts in the literature, the optimal scaling of expected error for this class of functions in the $m\gg n$ regime is still an open problem. 




	\section*{Acknowledgments}
	This research was supported by Iran National Science Foundation (INSF) under contract No. 970128.

\bibliography{ref}
\bibliographystyle{plainnat}

\newpage

\medskip
\medskip
\section*{\centering \huge Appendices }
\appendix


\section{Proof of Proposition~\ref{prop:H2constb}} \label{app:constant bit upper bound}
For simplicity, in this proof we will be working with the $[0,1]$ interval as the domain.
Consider the following randomized algorithm: 
\begin{itemize}
	\item Suppose that each machine $i$ observes $n$ function $f_1^i,\cdots,f_n^i$ and finds the minimizer  of $\sum_{j=1}^n f_j^i(\theta)$, which we denote by $\theta^i$. 
	Machine $i$ then sends a signal $Y^i$ of the following form
	\begin{equation*}
	Y^i=\begin{cases}
	0, \qquad\mbox{with prob.    } \theta^i,\\
	1, \qquad \mbox{with prob.    } 1-\theta^i.
	\end{cases}
	\end{equation*}
	\item The server receives signals from all machines and outputs $\hat{\theta}=1/m\sum_{i=1}^m Y^i$.
\end{itemize}
For the above algorithm, we have
\begin{equation} \label{eq:var of theta hat}
\operatorname{var}\big(\hat{\theta}\big)\,=\,\operatorname{var}\Bigg(\frac{1}{m} \sum_{i=1}^m Y^i\Bigg)
\,=\,\frac{1}{m}\operatorname{var}\big(Y^1\big)
\,=\,O\left(\frac{1}{m}\right),
\end{equation}
where the last equality is because $Y^1$ is a binary random variable. 
Then,
\begin{align*}
\mathbb{E}\Big[\big(\hat{\theta}-\theta^*\big)^2\Big]\,&=\,\mathbb{E}\Big[\big(\hat{\theta}-\mathbb{E}[\hat{\theta}]+\mathbb{E}[\hat{\theta}]-\theta^*\big)^2\Big]\\
&=\, \mathbb{E}\Big[\big(\hat{\theta}-\mathbb{E}[\hat{\theta}]\big)^2\Big]
+\mathbb{E}\Big[\big(\mathbb{E}[\hat{\theta}]-\theta^*\big)^2\Big]\\
&=\,\operatorname{var}\big(\hat{\theta}\big)+\Big(\mathbb{E}\big[\hat{\theta}-\theta^*\big]\Big)^2\\
&=\, O\left(\frac{1}{m}\right)+O\left(\frac{1}{n}\right),
\end{align*}
where the last equality is due to \eqref{eq:var of theta hat} and Lemma~\ref{lemma:10}. 
This completes the proof of Proposition~\ref{prop:H2constb}.

\medskip
\section{Proof of Proposition~\ref{lem:naive}} \label{app:proof naive}
For the ease of notation, we denote the number of grid points by $k=\sqrt[3]{m}/\log(m)$. Then, the probability of choosing each point $\theta$ in the grid  equals $1/k$. 
Let $N(\theta)$ be the number of machines that select grid point $\theta$. From Hoeffding's inequality (cf. Lemma~\ref{lemma:CI}~(a)), for any grid point $\theta$,
\begin{align*}
\Pr\left(N(\theta)<\frac{m}{2k}\right)\,&=\,\Pr\left(-N(\theta)+\frac{m}{k}>\frac{m}{2k}\right)\\
&\le \,\Pr\left(\frac1m \Big|N(\theta)-\E\big[N(\theta)\big]\Big| >\frac{1}{2k}\right)\\
&\leq\, \exp\left(-2m\,\left(\frac{1}{2k}\right)^2\right)\\
&=\,\exp\left(\frac{-\sqrt[3]{m}\log^2m}{2}\right).
\end{align*}
It then follows from union bound that  with probability at most 
$1-k\exp\big(-\Omega(\sqrt[3]{m})\Big)=1-\exp\big(-\Omega(\sqrt[3]{m})\Big)$, for any grid point $\theta$, we have $N(\theta)\ge {m}/(2k)$.
For the rest of the proof, we assume that $N(\theta)\ge {m}/(2k)$, for all grid points $\theta$.

Recall that the derivatives of the functions in $\F$ are assumed to be in the $[-1,1]$ interval (cf. Assumption~\ref{ass:1}).
Then, it follows from the Hoeffding's inequality (cf. Lemma~\ref{lemma:CI}~(a)) that for any point $\theta$ in the grid and any $\alpha\geq 1$,
\begin{equation*}
\Pr\left(\big|\hat{F}'(\theta)- F'(\theta)
\big|>\frac{\alpha}{k}\right)
\,\leq\, 2\exp\left(\frac{-2}{2^2}\times\frac{m}{2k}\times \left(\frac{\alpha}{k}\right)^2\right)
\,=\, 2\exp\left(-\frac{\alpha^2\log^2m}4\right). 
\end{equation*} 
Let $\mathcal{E}_{\alpha}$ be the event that for all grid points $\theta$, $\big|\hat{F}'(\theta)-F'(\theta)\big| \leq\alpha/k$. From  union bound, we have
\begin{equation}
\begin{split}
1-\Pr(\mathcal{E}_{\alpha})\,&\leq\,  2k\exp\left(-\frac{\alpha^2\log^3m}4\right)\,
=\, O\Big(\exp\big(-\alpha^2\log^3m\big)\Big),
\end{split}
\label{eq:5}
\end{equation}
where the equality is because $\alpha\ge 1$.
Let $\tilde{\theta}$ be the closest grid point to $\theta^*$. 
Then, $|\tilde\theta-\theta^*|\le 1/k$ and
it follows from the Lipschitz continuity of $F'$ that 
\begin{equation}\label{eq:F' theta tilde}
\big| {F}'(\tilde\theta) \big| \,=\,\big| {F}'(\tilde\theta)  -F'(\theta^*)\big| 
\,\le |\tilde\theta-\theta^*|\,\le\, \frac1k.
\end{equation}
Since $F$ is $\lambda$-strongly convex, for any $\alpha\ge 1$, we have
\begin{align*}
\Pr\left( \big| \hat{\theta} -\theta^*  \big| > \frac{3\alpha}{k\lambda} \right)\,
&\le\, \Pr\left( \big| F'(\hat{\theta}) -F'(\theta^*)  \big| > \frac{3\alpha}{k} \right)\\
&=\,\Pr\left( \big| F'(\hat{\theta}) \big| > \frac{3\alpha}{k} \right)\\
&\le\, \Pr\left( \big| F'(\hat{\theta}) -\hat{F}'(\hat\theta)  \big| > \frac{\alpha}{k} \right) 
\,+\, \Pr\left( \big| \hat{F}'(\hat{\theta}) \big| > \frac{2\alpha}{k} \right)\\
&\le^a\, O\Big(\exp\big(-\alpha^2\log^3m\big)\Big)
\,+\, \Pr\left( \big| \hat{F}'(\hat{\theta}) \big| > \frac{2\alpha}{k} \right)\\
&\le^b\, O\Big(\exp\big(-\alpha^2\log^3m\big)\Big)
\,+\, \Pr\left( \big| \hat{F}'(\tilde{\theta}) \big| > \frac{2\alpha}{k} \right)\\
&\le\, O\Big(\exp\big(-\alpha^2\log^3m\big)\Big)
\,+\, \Pr\left( \big| \hat{F}'(\tilde{\theta}) - {F}'(\tilde{\theta}) \big| > \frac{\alpha}{k} \right)
\,+\,\Pr\left( \big| {F}'(\tilde{\theta}) \big| > \frac{\alpha}{k} \right)\\
&\le^c\, O\Big(\exp\big(-\alpha^2\log^3m\big)\Big)\,+\, O\Big(\exp\big(-\alpha^2\log^3m\big)\Big) \,+\, 0\\
&=\, O\Big(\exp\big(-\alpha^2\log^3m\big)\Big),
\end{align*}
where $(a)$ follows from \eqref{eq:5}, $(b)$ is because $\hat\theta$ minimizes $ \big| \hat{F}'({\theta}) \big|$ over all grid points $\theta$, and $(c)$ is due to \eqref{eq:5} and \eqref{eq:F' theta tilde}.
This completes the proof of Proposition~\ref{lem:naive}.

\medskip
\section{Proof of Theorem~\ref{th:main upper}} \label{sec:proof main alg}
We first show that $s^*$ is a closest grid point of $G$ to $\theta^*$ with high probability. 
We then show that for any $l\le t$ and any $p\in \tilde{G}_{s^*}^l$, the number of received signals corresponding to $p$ is large enough so that the server obtains a good approximation of $\nabla F$ at $p$. 
Once we have a good approximation of $\nabla F$ at all points of $\tilde{G}_{s^*}^t$, a point with the minimum norm for this approximation lies close to the minimizer of $F$.

Suppose that machine $i$ observes functions $f_1^i,\ldots,f_n^i$. 
Recall the definition of $\theta^i$ in \eqref{eq:def theta i half upper}.
The following lemma provides a bound on $\theta^i-\theta^*$, which improves upon the bound in Lemma~8 of \citep{zhang2013information}. 

\begin{lemma}
	For any $i\le m$,
	\begin{equation*}
	\Pr\left(\|\theta^i-\theta^*\|\geq \frac{\alpha}{\sqrt{n}}\right)
	\,\leq\, d\exp\left(\frac{-\alpha^2 \lambda^2}{d}\right),
	\end{equation*}
	where $\lambda$ is the lower bound on the curvature of $F$ (cf. Assumption~\ref{ass:1}).
	\label{lemma:10}
\end{lemma}
The proof relies on concentration inequalities, and is given in Appendix~\ref{app:proof lem EM1}.
We collect two well-known concentration inequalities in the following lemma.
\begin{lemma} (Concentration inequalities)
	
	\begin{enumerate}
		\item[(a)] (Hoeffding's inequality)  
		Let $X_1,\cdots,X_n$ be independent random variables ranging over the interval $[a,a+\gamma]$. Let $\bar{X}=\sum_{i=1}^n X_i/n$ and $\mu =\mathbb{E}[\bar{X}]$.
		Then, for any $\alpha>0$,
		\begin{equation*}
		\Pr\big(|\bar{X}-\mu|>\alpha\big)\leq 2\exp\left(\frac{-2n\alpha^2}{\gamma^2}\right).
		\end{equation*}
		
		\item[(b)] (Theorem 4.2 in \cite{motwani1995randomized})  
		Let $X_1,\cdots,X_n$ be independent Bernoulli  random variables, ${X}=\sum_{i=1}^n X_i$, and $\mu =\mathbb{E}[{X}]$.
		Then, for any $\alpha\in(0,1]$,
		\begin{equation*}
		\Pr\big(X<(1-\alpha)\mu\big)\leq \exp\left(-\frac{\mu\alpha^2}{2}\right).
		\end{equation*}
	\end{enumerate}
	\label{lemma:CI}
\end{lemma}

Based on the above lemma, we have
\begin{equation}\label{eq:union for theta i}
\begin{split}
\Pr\bigg(\|\theta^i-\theta^*\|&\leq \frac{\log(mn)}{2\sqrt{n}},\, \mbox{for } i=1,\ldots,m\bigg)\\ &\geq\,  1-m \Pr\left(\|\theta^1-\theta^*\|\geq \frac{\log(mn)}{2\sqrt{n}}\right)\\
&\geq 1-md \exp\left(\frac{-\lambda^2\log^2(mn)}{4d}\right) \\
&=\,1-\exp\Big(-\Omega\big(\log^2(mn)\big)\Big),
\end{split}
\end{equation}
where the first inequality is due to the union bound and the fact that the  distributions of $\theta^1,\ldots,\theta^m$ are identical, and the second inequality follows from Lemma~\ref{lemma:10}.
Thus, with probability at least $1-\exp\big(-\Omega(\log^2(mn))\big)$, every $\theta^i$ is  in the distance $\log(mn)/2\sqrt{n}$ from $\theta^*$. 
For each machine $i$, let $s^i$ be the $s$-component of machine $i$'s signal.
Therefore, with probability at least $1-\exp\big(-\Omega(\log^2(mn))\big)$, for any machine $i$,
\begin{align*}
\Pr\left(\|s^i-\theta^*\|_\infty > \frac{\log(mn)}{\sqrt{n}} \right)\,
&\le\, \Pr\left(\|s^i-\theta^i\|_\infty
+ \|\theta^i-\theta^*\|_\infty > \frac{\log(mn)}{\sqrt{n}} \right)\\
&\le\, \Pr\left(\|s^i-\theta^i\|_\infty  > \frac{\log(mn)}{2\sqrt{n}} \right) \\
&\qquad + \Pr\left( \|\theta^i-\theta^*\|_\infty > \frac{\log(mn)}{2\sqrt{n}} \right)\\
& = \, 0 +   \Pr\left( \|\theta^i-\theta^*\|_\infty > \frac{\log(mn)}{2\sqrt{n}} \right)\\
& = \,   \exp\Big(-\Omega\big(\log^2(mn)\big)\Big),
\end{align*}
where the first equality is due to the choice of  $s^i$ as the nearest grid point to $\theta^i$, and the last equality follows from \eqref{eq:union for theta i}.
Recall that $s^*$ is the grid point with the largest number of occurrences in the received signals.
Therefore, with probability at least $1-\exp\big(-\Omega(\log^2(mn))\big)$,
\begin{equation} \label{eq:s* in D}
\|s^*-\theta^*\|_\infty \leq \frac{\log(mn)}{\sqrt{n}};
\end{equation}
equivalently, $\theta^*$ lies in the $\big(2\log(mn)/\sqrt{n}\big)$-cube $C_{s^*}$ centered at $s^*$. 

Let $m^*$ be the number of machines that select $s=s^*$. 
We let $\mathcal{E'}$  be the event that $m^*\geq m/2^d$.
Since the grid $G$ has block size $2\log(mn)/\sqrt{n}$, there are at most $2^d$ points $s$ of the grid that satisfy
$\|s-\theta^*\|_\infty \leq \log(mn)/\sqrt{n}$.
It then follows from \eqref{eq:s* in D} that
\begin{equation}\label{eq:prb E1 upper}
\Pr\big(\mathcal{E'}\big)=1-\exp\big(-\Omega(\log^2(mn))\big).
\end{equation}



We now turn our focus to the inside of cube $C_{s^*}$.
Let 
\begin{equation} \label{eq:def eps upper}
\epsilon \,\triangleq\, \frac{2\sqrt{d}\,\log(mn)}{\sqrt{n}}\times \delta \,=\,  \frac{8d \log^{1+\frac{5}{\max(d,2)}}(mn)}{n^{\frac12}\, m^{\frac{1}{\max(d,2)}}}.
\end{equation}
For any $p\in \bigcup_{l\le t} \tilde{G}_{s^*}^l$, let $N_p$ be the number of machines that select point $p$.
Let $\mathcal{E''}$ be the event that for any $l\le t$ and any $p\in \tilde{G}_{s^*}^l$, we have 
\begin{equation} \label{eq:bound Np}
N_p \ge \frac{d^22^{-2l}\log^6(mn)}{2n\epsilon^2}.
\end{equation}
Then,
\begin{lemma} \label{lem:prob of E2 upper}
	$\Pr\big(\mathcal{E''} \big) = 1-\exp\big(-\Omega(\log^2(mn))\big)$.
\end{lemma}
The proof is based on the concentration inequality in Lemma~\ref{lemma:CI}~(b), and is given in Appendix~\ref{app:proof lem E2 upper}.

Capitalizing on Lemma~\ref{lem:prob of E2 upper}, we now obtain a bound on the estimation error of gradient of $F$ at the grid points in $\tilde{G}_{s^*}^l$. 
Let $\mathcal{E'''}$ be the event that for any $l\le t$ and any grid point $p\in \tilde{G}_{s^*}^l$, we have $$\big\|\hat{\nabla} F(p) -\nabla F(p)\big\|\,<\,\frac{\epsilon}{4}.$$

\begin{lemma} \label{lem:prob E3 upper}
	$\Pr\big(\mathcal{E'''}\big)\,=\,1-\exp\big(-\Omega(\log^2(mn))\big)$. 
\end{lemma}
The proof is given in Appendix~\ref{app:proof E3 upper} and relies on Hoeffding's inequality and the lower  bound on the number of received signals for each grid point, driven in Lemma~\ref{lem:prob of E2 upper}.

In the remainder of the proof, we assume that \eqref{eq:s* in D} and  $\mathcal{E'''}$ hold. 
Let $p^*$ be the  closest grid point in $\tilde{G}_{s^*}^t$ to $\theta^*$.
Therefore, 
\begin{equation}\label{eq:p theta eps2}
\|p^*-\theta^*\|\,\leq\, \sqrt{d}\,2^{-t}\frac{\log(mn)}{\sqrt{n}} \,=\,  \epsilon/2.
\end{equation}
Then, it follows from $\mathcal{E'''}$ that
\begin{equation}
\begin{split}
\|\hat{\nabla} F(p^*)\|&\leq \big\|\hat{\nabla} F(p^*)-\nabla F(p^*)\big\|\,+\,\|\nabla F(p^*)\|\\
&\leq\, \epsilon/4+\|\nabla F(p^*)\|\\
&=\,\epsilon/4+ \big\|\nabla F(p^*)-\nabla F(\theta^*)\big\|\\
&\leq\, \epsilon/4 +\big\|p^*-\theta^*\big\|\\
&\leq\, \epsilon/4+\epsilon/2\\
&=\, 3\epsilon/4,
\end{split}
\label{eq:11star}
\end{equation}
where the second inequality is due to $\mathcal{E'''}$, the third inequality follows from the Lipschitz continuity of $\nabla F$, and the last inequality is from \eqref{eq:p theta eps2}.
Therefore, 
\begin{align*}
\|\hat{\theta}-\theta^*\|\,&\leq\, \frac{1}{\lambda}\,\big\|\nabla F(\hat{\theta})-\nabla F(\theta^*)\big\|\\
&=\,\frac{1}{\lambda}\,\|\nabla F(\hat{\theta})\|\\
&\leq \,\frac{1}{\lambda}\,\|\hat{\nabla} F(\hat{\theta})\|+\frac{1}{\lambda}\,\big\|\hat{\nabla} F(\hat{\theta})-\nabla F(\hat{\theta})\big\|\\
&\leq^a\, \frac{1}{\lambda}\,\|\hat{\nabla} F(\hat{\theta})\|+\frac{\epsilon}{4\lambda}\\
& \leq^b\, \frac{1}{\lambda}\,\|\hat{\nabla} F(p^*)\|+\frac{\epsilon}{4\lambda}\\
&\leq^c\, \frac{3\epsilon}{4\lambda}+\frac{\epsilon}{4\lambda}\\
&=\,\frac{\epsilon}{\lambda},
\end{align*}
($a$) Due to event $\mathcal{E'''}$. 
\\ ($b$) Because the output of the server, $\hat{\theta}$, is a grid point $p$ in $\tilde{G}_{s^*}^t$  with  smallest $\|\hat{\nabla}F(p)\|$.
\\ ($c$) According to \eqref{eq:11star}.
\\ Finally, it follows from \eqref{eq:s* in D} and Lemma~\ref{lem:prob E3 upper} that the above inequality holds with probability $1-\exp\big(-\Omega (\log^2(mn))\big)$. 
Equivalently,
\begin{equation*}
\Pr\left(\|\hat{\theta}-\theta^*\| \ge \frac{\epsilon}{\lambda} \right)\,=\,\exp\Big(-\Omega\big(\log^2(mn)\big)\Big),
\end{equation*}
and Theorem~\ref{th:main upper} follows.


\medskip
\section{Proof of Lemma~\ref{lemma:10}} \label{app:proof lem EM1}
Let $F^i(\theta)=\sum_{j=1}^{n/2} f_j^i(\theta)$, for all $\theta\in[-1,1]^d$.
From the lower bound $\lambda$ on the second derivative of $F$, we have  
$$\| \nabla F(\theta^i) - \nabla {F}^i(\theta^i)\| \,=\, \|\nabla F(\theta^i)\| \,=\, \|\nabla F(\theta^i) - \nabla F(\theta^*) \| \,\geq\, \lambda \|\theta^i-\theta^*\|, $$
where  the two equalities are because $\theta^i$ and $\theta^*$ are the the minimizers of ${F}^i $ and $F$, respectively.  
Then,
\begin{equation}
\begin{split}
\Pr\left(\|\theta^i-\theta^*\| \ge \frac{\alpha}{\sqrt{n}}\right) \,& \le \, 
\Pr\left(\| \nabla F(\theta^i) - \nabla {F}^i(\theta^i)\| \geq \frac{\lambda\alpha}{\sqrt{n}}\right)\\
& \le^{a}\, \sum_{j=1}^d \Pr\left(\Big|\frac{\partial {F}^i(\theta^i)}{\partial \theta_j} - \frac{\partial {F}(\theta^i) }{\partial \theta_j}\Big| > \, \frac{\alpha \lambda}{\sqrt{d}\sqrt{n}}\right)\\
& =\, d\,\Pr\left(\Big|\frac{2}{n}\sum_{l=1}^{n/2} \frac{\partial}{\partial \theta_j} f_l^i(\theta^i) \,-\,   \E_{f \sim P}\big[\frac{\partial}{\partial \theta_j} f(\theta^i) \big] \Big|\geq \frac{\alpha \lambda}{\sqrt{d}\sqrt{n}} \right)\\
& =^b \, d \exp\left(-\frac{\alpha^2 \lambda^2}{d}\right),
\end{split}
\end{equation}
\\($a$) Follows from the union bound and the fact that for any $d$-dimensional vector $v$, there exists an entry $v_i$ such that $\|v\|\le |v_i|/\sqrt{d}$.
\\($b$) Due to Hoeffding's inequality (cf. Lemma~\ref{lemma:CI}~(a)).\\
This completes the proof of Lemma~\ref{lemma:10}.


\section{Proof of Lemma~\ref{lem:prob of E2 upper}} \label{app:proof lem E2 upper}
Recall that for any $l\le t$, given $s = s^*$, the probability that $p\in \tilde{G}_{s^*}^l$ is $2^{(d-2)l}/\sum_{j=0}^t2^{(d-2)j}$.
Assuming $\mathcal{E'}$, for any $l\le t$ and any $p \in \tilde{G}_{s^*}^l$,
\begin{equation}
\begin{split}
\mathbb{E}\big[N_p\big]\,&=\, 2^{-dl} \times \frac{2^{(d-2)l}}{\sum_{j=0}^t2^{(d-2)j}} \times m^*\\
\,&=\, \frac{2^{-2l}m^*}{\sum_{j=0}^t2^{(d-2)j}}\\
&\geq^a\, \frac{2^{-2l}m}{2^d}\times \frac{1}{\sum_{j=0}^t 2^{(d-2)j}}\\
&\geq^b\, \frac{2^{-2l}m}{2^d}\times\frac{1}{t2^{t\max(0,d-2)}}\\
&=^c\,\frac{2^{-2l}m}{2^d}\times \frac{\delta^{\max(0,d-2)}}{\log(1/\delta)}\\
&\geq^d\, \frac{2^{-2l}m}{2^d\delta^2}\times \frac{\delta^{\max(d,2)}}{\log(mn)}\\
&=^e\, \frac{2^{-2l }m}{2^d \delta^2\log(mn)}\times (4\sqrt{d})^{\max(d,2)}\frac{\log^5(mn)}{m}\\
&\geq\, \frac{4d2^{-2l}\log^4(mn)}{\delta^2},
\end{split}
\label{eq:8star}
\end{equation}
where
($a$) follows from $\mathcal{E'}$,
($b$) is valid for all non-negative integers $t$ and $d$, 
($c$) is from the definition of $t=\log(1/\delta)$, 
($d$) is due to the fact that $1/\delta \le \sqrt{m} \leq mn$ 
and ($e$) is because of the definition of $\delta$ in \eqref{eq:def delta}.
Then,
\begin{equation}
\begin{split}
\Pr\left(N_p\leq \frac{d^2 2^{-2l}\log^6(mn)}{2n\epsilon^2}\right)\, &=\, 
\Pr\left(N_p\leq \frac12 \times \frac{d 2^{-2l}\log^4(mn)}{4\delta^2}\right)\\
&\leq \,\Pr\left(N_p\leq \frac{\mathbb{E}[N_p]}{2}\right)\\
& \leq \,2^{-(1/2)^2\mathbb{E}[N_p]/2}\\
&=\,\exp\big(-\Omega(\log^4(mn))\big),
\end{split}
\end{equation}
where the first equality is from the definition of $\epsilon$ in \eqref{eq:def eps upper}, the first inequality is due to \eqref{eq:8star}, the second inequality follows from Lemma~\ref{lemma:CI}~(b), and the last equality is due to  \eqref{eq:8star} and the fact that $2^{-2l_p}\geq 2^{-2t}=2^{-2\log(1/\delta)}=\delta^2$.
Then,
\begin{align*}
\Pr\big(\mathcal{E''} \mid \mathcal{E'}\big)\, &\geq\, 1- \sum_{l=0}^{t} \sum_{p\in \tilde{G}_{s^*}^l} \Pr\left(N_p\leq \frac{d^2 2^{-2l_p}\log^6(mn)}{2n\epsilon^2}\right)\\
&\geq\, 1- t2^{dt} \exp\Big(-\Omega\big(\log^4(mn)\big)\Big)\\
&\geq\, 1- \log(1/\delta)\left(\frac1\delta\right)^d \exp\Big(-\Omega\big(\log^4(mn)\big)\Big)\\
&>\,1-m\log(m)\exp\Big(-\Omega\big(\log^4(mn)\big)\Big)\\
&=\,1-\exp\Big(-\Omega\big(\log^4(mn)\big)\Big).
\end{align*}
On the other hand, we have from \eqref{eq:prb E1 upper} that $\Pr\big(\mathcal{E'} \big)= 1-\exp\big(-\Omega(\log^2(mn))\big)$.
Then, $\Pr\big(\mathcal{E''} \big)= 1-\exp\big(-\Omega(\log^2(mn))\big)$ and Lemma~\ref{lem:prob of E2 upper} follows.


\section{Proof of Lemma~\ref{lem:prob E3 upper}} \label{app:proof E3 upper}
For any $l\le t$ and any $p\in \tilde{G}_{s^*}^0$, let
$$\hat{\Delta}(p)= \frac{1}{N_p}\,\, \sum_{\substack{\mbox{\scriptsize{Signals of the form }} \\ Y^i=(s^*,p,\Delta)}}\Delta,$$
and let $\Delta^*(p)= \mathbb{E}[\hat{\Delta}(p)]$.

We first consider the case $l=0$. Note that $\tilde{G}_{s^*}^0$ consists of a single point $p=s^*$.
Moreover, the component $\Delta$ in each signal is the average over the gradient of $n/2$ independent functions. 
\hide{The problem with the correlation of $s$ and $\Delta$. Note that the distribution of functions in a machine that chooses $s$ is not the same as $P$. Use $n/2$ of the functions for choosing $s$ and $n/2$ for evaluating $\Delta$ at $p$. $n/2$ or $log n$ for $s$?}
Then, $\hat{\Delta}(p)$ is the average over the gradient of $N_p \times n/2$ independent functions.
Given event $\mathcal{E''}$, for any entry $j$ of the gradient, it follows from Hoeffding's inequality (Lemma~\ref{lemma:CI}~(a)) that
\begin{equation}
\begin{split}
\Pr&\left(\big|\hat{\Delta}_j(p)-\Delta^*_j(p)\big|\geq \frac{\epsilon}{4\sqrt{d}\log(mn)}\right)\\
&\qquad\leq\, \exp\left(-N_pn \times \left(\frac{\epsilon}{4\sqrt{d}\log(mn)}\right)^2 \,/\,2^2 \right)\\
&\qquad\leq \exp\left(-n \frac{d^2\log^6(mn)}{8n\epsilon^2} \times\frac{\epsilon^2}{16 d\log^2(mn)}\right)\\
&\qquad=\exp\left(\frac{-d\log^4(mn)}{128}\right)\\
&\qquad=\, \exp\Big(-\Omega\big(\log^4(mn)\big)\Big).
\end{split}
\label{eq:10star}
\end{equation}

For $l\geq 1$, consider a grid point $p\in \tilde{G}_{s^*}^l$ and let $p'$ be the parent of $p$.
Then, $\|p-p'\|={\sqrt{d}\,2^{-l}\log(mn)}/\sqrt{n}$.
Furthermore, for any function $f\in \F$, we have $\|\nabla f(p)-\nabla f(p')\|\leq \|p-p'\|$. Hence, for any $j\le n$,
\begin{equation*}
\Big|\frac{\partial f(p)}{\partial x_j}-\frac{\partial f(p')}{\partial x_j}\Big|
\,\leq\, \|p-p'\|
\,=\,\frac{\sqrt{d}\log(mn)2^{-l}}{\sqrt{n}}.
\end{equation*}
Therefore, $\hat{\Delta}_j(p)$ is the average of $N_p \times n/2$ independent variables with absolute values no larger than $\gamma\triangleq {\sqrt{d}\log(mn)2^{-l}}/{\sqrt{n}}$. 
Given event $\mathcal{E''}$, it then follows from the Hoeffding's inequality that
\begin{align*}
\Pr&\left(\big|\hat{\Delta}_j(p)-\Delta_j^*(p)\big|
\geq\, \frac{\epsilon}{4\sqrt{d}\log(mn)}\right)\\
&\leq\, \exp\left(-{nN_p}\times\frac1{(2\gamma)^2}\times \left(\frac{\epsilon}{4\sqrt{d}\log(mn)}\right)^2\right)\\
&\leq^a\, \exp\left(-n \frac{d^2 2^{-2l}\log^6(mn)}{2n\epsilon^2}\times \frac{n}{4d2^{-2l}\log^2(mn)}\times \frac{\epsilon^2}{16 d \log^2(mn)}\right)\\
&=\,\exp\big(-n\log^2(mn)/128\big)\\
&=\,\exp\Big(-\Omega\big(\log^2(mn)\big)\Big),
\end{align*}
where the second inequality is by substituting $N_p$ from \eqref{eq:bound Np}. 
Employing union bound, we obtain
\begin{align*}
\Pr&\left(\big\|\hat{\Delta}(p)-\Delta^*(p)\big\| \ge \frac{\epsilon}{4\log(mn)}\right)\\
&\qquad\leq\, \sum_{j=1}^d\Pr\left(\big|\hat{\Delta}_j(p)-\Delta_j^*(p)\big|\geq \frac{\epsilon}{4\sqrt{d}\log(mn)}\right)\\
&\qquad =\, d \, \exp\Big(-\Omega\big(\log^2(mn)\big)\Big)\\
&\qquad=\exp\Big(-\Omega\big(\log^2(mn)\big)\Big).
\end{align*}

Recall from \eqref{eq:page17} that for any non-zero $l\leq t$ and any $p \in \tilde{G}_{s^*}^l$ with parent $p'$,
$$\hat{\nabla} F(p)- \nabla F(p)\,=\,\hat{\nabla} F(p')-\nabla F(p')+\hat{\Delta}(p)-\Delta^*(p).$$
Then,
\begin{align*}
\Pr&\left(\big\|\hat{\nabla} F(p)-\nabla F(p)\big\|>\frac{(l+1)\epsilon}{4\log(mn)}\right)\\
&\leq\,\Pr\left(\big\|\hat{\nabla} F(p')-\nabla F(p')\big\|>\frac{l\epsilon}{4\log(mn)}\right)\\
&\qquad +\Pr\left(\big\|\hat{\Delta}(p)-\Delta^*(p)\big\|>\frac{\epsilon}{4\log(mn)}\right)\\
&\leq\,\Pr\left(\big\|\hat{\nabla} F(p')-\nabla F(p')\big\|>\frac{l\epsilon}{4\log(mn)}\right) +\exp\Big(-\Omega\big(\log^2(mn)\big)\Big).
\end{align*}
Employing an induction on $l$, we obtain for any $l\le t$,
$$\Pr\left(\big\|\hat{\nabla} F(p)-\nabla F(p)\big\|>\frac{(l+1)\epsilon}{4\log(mn)}\right)
\,\le\,\exp\Big(-\Omega\big(\log^2(mn)\big)\Big).$$
Therefore, for any grid point $p$, 
\begin{align*}
\Pr\left(\big\|\hat{\nabla} F(p)-\nabla F(p)\big\|>\frac{\epsilon}{4}\right)\,
&\le\,\Pr\left(\big\|\hat{\nabla} F(p)-\nabla F(p)\big\|>\frac{(t+1)\epsilon}{4\log(mn)}\right)\\
&=\exp\Big(-\Omega\big(\log^2(mn)\big)\Big),
\end{align*} 
where the inequality is because $t+1 = log(1/\delta)+1\le \log(mn)$.
It then follows from the union bound that
\begin{equation*}
\begin{split}
\Pr\big(\mathcal{E'''} \mid \mathcal{E''}\big)\, &\geq\,  1- \sum_{l=0}^{t} \sum_{p\in \tilde{G}_{s^*}^l}\Pr\left(\big\|\hat{\nabla} F(p)-\nabla F(p)\big\|>\frac{\epsilon}{4}\right)\\
&\geq\, 1- t2^{dt} \exp\Big(-\Omega\big(\log^2(mn)\big)\Big)\\
&=\, 1- \log(1/\delta)\left(\frac1\delta\right)^d \exp\Big(-\Omega\big(\log^2(mn)\big)\Big)\\
&\ge \,1-m\log(m)\exp\Big(-\Omega\big(\log^2(mn)\big)\Big)\\
&=\,1- \exp\Big(-\Omega\big(\log^2(mn)\big)\Big).
\end{split}
\end{equation*} 
On the other hand, we have from Lemma~\ref{lem:prob of E2 upper} that $\Pr\big(\mathcal{E''} \big)= 1-\exp\big(-\Omega(\log^2(mn))\big)$.
Then, $\Pr\big(\mathcal{E'''} \big)= 1-\exp\big(-\Omega(\log^2(mn))\big)$ and Lemma~\ref{lem:prob E3 upper} follows.

\end{document}